\documentclass[a4paper]{article}

\usepackage{INTERSPEECH2021}
\usepackage{multirow}
\usepackage{cite}
\usepackage{hyperref} 
\usepackage{xcolor}  

\title{Towards Lifelong Learning of End-to-end ASR}
\name{Heng-Jui Chang, Hung-yi Lee, Lin-shan Lee}
\address{
  School of Electrical Engineering and Computer Science, National Taiwan University, Taiwan
 }
\email{\{b06901020, hungyilee\}@ntu.edu.tw, lslee@gate.sinica.edu.tw}

\begin{document}

\maketitle
\begin{abstract}
    Automatic speech recognition (ASR) technologies today are primarily optimized for given datasets; thus, any changes in the application environment (e.g., acoustic conditions or topic domains) may inevitably degrade the performance.
    We can collect new data describing the new environment and fine-tune the system, but this naturally leads to higher error rates for the earlier datasets, referred to as catastrophic forgetting.
    The concept of lifelong learning (LLL) aiming to enable a machine to sequentially learn new tasks from new datasets describing the changing real world without forgetting the previously learned knowledge is thus brought to attention.
    This paper reports, to our knowledge, the first effort to extensively consider and analyze the use of various approaches of LLL in end-to-end (E2E) ASR, including proposing novel methods in saving data for past domains to mitigate the catastrophic forgetting problem.
    An overall relative reduction of 28.7\% in WER was achieved compared to the fine-tuning baseline when sequentially learning on three very different benchmark corpora.
    This can be the first step toward the highly desired ASR technologies capable of synchronizing with the continuously changing real world.
\end{abstract}

\section{Introduction}
\label{sec:intro}

The real world is changing and evolving from time to time, and therefore machines naturally need to update and adapt to the new data they receive.
However, when a trained deep neural network was adapted to a new dataset with a different distribution, it often loses the knowledge previously acquired and performs the previous task worse than before.
This phenomenon is called \textit{catastrophic forgetting} \cite{Mccloskey89-catastrophic}.
Under this scenario, people try to re-train the models from scratch with both the past and the new data jointly, sometimes referred to as \textit{multitask learning}.
For various reasons, including privacy issues and the limited storage capacity, the earlier data are unlikely to be kept forever.
Therefore, \textit{lifelong learning} (LLL) or \textit{continual learning} \cite{Chen18-llml}, aiming for training a single model to perform a stream of tasks without forgetting those learned earlier, not relying on keeping all training data from the beginning, becomes a necessary goal for the continuously changing real world.

In general, LLL approaches can be categorized into three types.
Regularization-based methods aim to consolidate essential parameters in a model by adding regularization terms in the loss function \cite{Kirkpatrick17-ewc, Schwarz18-online-ewc, Zenke17-si, Aljundi18-mas, Ehret20-rnn-hypernet}.
Architecture-based methods try to assign some model capacity for each task or expand the model size to handle additional tasks \cite{Rusu16-progressive, Fernando17-pathnet, Mallya18-packnet}.
Data-based methods then try to save or generate some samples from the past tasks to prevent catastrophic forgetting \cite{Lopez17-gem, Li17-lwf, Sun19-lamol, Von19-hnet, Saha21-gpm}.
Studies of LLL have been reported more on computer vision \cite{Kirkpatrick17-ewc, Zenke17-si, Aljundi18-mas, Ehret20-rnn-hypernet,Fernando17-pathnet, Mallya18-packnet, Lopez17-gem, Li17-lwf, Von19-hypernet, Benavides20-knowlll, Mendez20-compositional, Wen20-batchensemble, Von19-hnet, Saha21-gpm} and reinforcement learning \cite{Kirkpatrick17-ewc, Schwarz18-online-ewc, Rusu16-progressive, Fernando17-pathnet, Rostami20-task-descript}, yet much less on automatic speech recognition (ASR) tasks \cite{Xue19-mtl-asr, Sadhu20-continual, Houston20-continual, Fu20-incremental}.

ASR technologies are very successful globally, and end-to-end (E2E) ASR approaches \cite{Graves14-ctcasr, Graves12-rnnt, Chan16-las, Gulati20-conformer} are very powerful in recent years, but with performance inevitably degraded in almost all cases over data disparate from training sets, e.g., in acoustic conditions or topic domains.
Various domain adaptation approaches for ASR were shown successful \cite{Hsu17-unsupervised-adapt, Meng17-unsupervised-adapt, Winata20-adapt-accent} in new domains, although inevitably suffering from serious catastrophic forgetting \cite{Xue19-mtl-asr, Sadhu20-continual, Houston20-continual}.
As a result, ASR technologies today remain unable to evolve with the changing real world.

To our knowledge, this is the first paper extensively studying the various concepts of LLL applied to E2E ASR.
Sadhu and Hermansky \cite{Sadhu20-continual} used model expansion for LLL on HMM-DNN ASR.
Houston and Kirchhoff \cite{Houston20-continual} used regularization methods for multi-dialect acoustic models.
However, E2E ASR discussed here jointly considers acoustic and language modeling in a single network and is thus different and challenging.
We compare and analyze regularization- and data-based methods and found the latter very effective, including proposing data selection approaches based on either perplexity or utterance duration.
Evaluation of CTC \cite{Graves14-ctcasr} ASR on WSJ \cite{Paul92-WSJ}, LibriSpeech \cite{Panayotov15-libri}, and Switchboard \cite{Godfrey92-swb} showed an overall relative WER reduction of 28.7\% compared to the fine-tuning baseline.

\section{Methods}
\label{sec:method}

\begin{figure}[t]
	\centering
	\includegraphics[width=\linewidth]{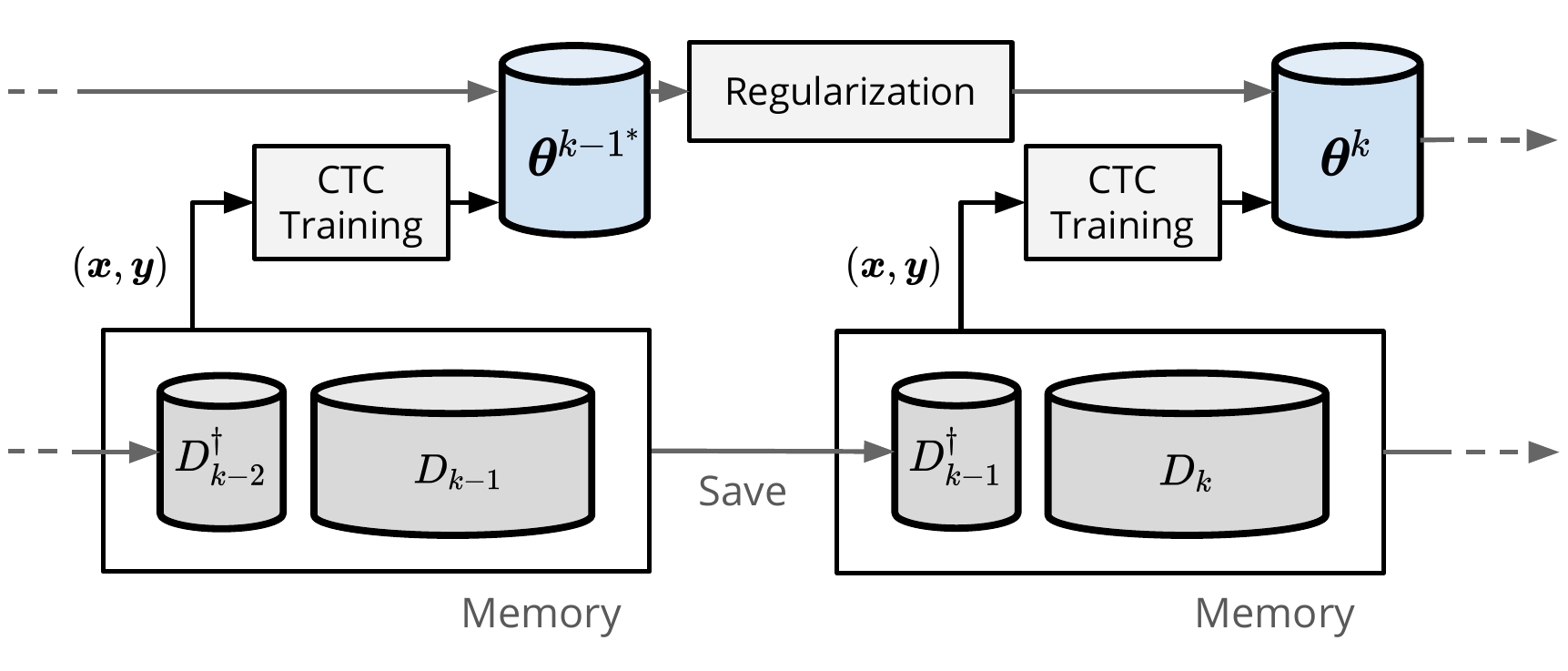}
    \vspace{-22pt}
	\caption{The training framework for LLL for E2E ASR.
	The $k^{\mathrm{th}}$ model ($\boldsymbol{\theta}^k$) is being trained on corpus $D_k$ and can reuse some limited past samples stored in $D_{k-1}^{\dagger}$.
	Also, the model trained in the previous stage ($\boldsymbol{\theta}^{k-1^\ast}$) can be used for regularization.
	}
	\label{fig:framework}
    \vspace{-5pt}
\end{figure}
\subsection{General Training Framework}
\label{subsec:framework}

Consider a training framework as in Fig. \ref{fig:framework} with $K$ training corpora $D_1$ to $D_K$ from different domains.
The E2E ASR (CTC \cite{Graves14-ctcasr} in this paper) is first trained with $D_1$, with parameters obtained denoted as $\boldsymbol{\theta}^{1^\ast}$, where "$\ast$" indicates the best parameter set.
The model is then trained in each stage ($k^{\mathrm{th}}$ stage in the right part of Fig. \ref{fig:framework}) on one corpus ($D_k$) at a time, and is allowed to reuse some samples from the previous corpora ($D^{\dagger}_{k-1}$) stored in a memory with a fixed capacity.
The target of LLL is to preserve high recognition accuracy for previous domains, or the last CTC model ($\boldsymbol{\theta}^{K^\ast}$) perform well on all domains $D_1$ to $D_K$.
All training data are transcribed audio-text pairs $(\boldsymbol{x}, \boldsymbol{y})$.
The $i^{\mathrm{th}}$ parameter of the CTC model trained on $D_k$ is $\theta_{i}^k$.

\subsection{Regularization-based Methods}
\label{subsec:method-reg}

\noindent\textbf{Elastic Weight Consolidation (EWC).}
We adopt the EWC \cite{Kirkpatrick17-ewc} and online EWC algorithms \cite{Schwarz18-online-ewc} previously proposed.
A regularization term is used to constrain parameters to stay close to those for previous tasks.
The loss function can be written as
\begin{equation}
    \mathcal{L}(\boldsymbol{\theta}^k) = \mathcal{L}_{\mathrm{CTC}}(\boldsymbol{\theta}^k) + \frac{\lambda}{2}\sum_i \Omega^k_i(\theta_{i}^{k}-\theta_{i}^{k-1^\ast})^2,
    \label{eq:ewc}
\end{equation}
where $\mathcal{L}_{\mathrm{CTC}}$ is the CTC loss, and $\Omega^k_i$ for the importance of each parameter is the diagonal of the Fisher information matrix.

\noindent\textbf{Synaptic Intelligence (SI).}
We also adopt SI \cite{Zenke17-si}, which is similar to EWC, but the importance measure $\Omega^k_i$ in Eq. (\ref{eq:ewc}) is estimated iteratively by its contribution to decreasing the loss,
\begin{equation}
    \Omega^k_i = \Omega^{k-1}_i + \frac{\omega^{k-1}_i}{\left(\theta_{i}^{k-1^\ast} - \theta_{i}^{k-2^\ast}\right)^2 + \xi},
    \label{eq:si_accum}
\end{equation}
where $\omega^{k-1}_i$ is obtained with the gradient of CTC loss and $\xi$ is a small constant to stabilize training.

\noindent\textbf{Knowledge Distillation (KD).}
Instead of limiting the parameters directly as done previously, KD minimizes the KL divergence between the output distributions of the current model ($\boldsymbol{\theta}^k$) and the previous model ($\boldsymbol{\theta}^{k-1^\ast}$) \cite{Xue19-mtl-asr, Fu20-incremental, Hinton15-knowledge-distill, Yu13-kl, Li17-lwf}.
The KD loss function can be written as
\begin{equation}
    \mathcal{L}_{\mathrm{KD}} = \mathrm{KL} \left[ \boldsymbol{y}'_{k-1} || \boldsymbol{y}'_{k} \right],
    \label{eq:kl_loss}
\end{equation}
where $\boldsymbol{y}'_{k-1}$ and $\boldsymbol{y}'_k$ are the output probability sequences for the previous and the current models $\boldsymbol{\theta}^{k-1^\ast}$ and $\boldsymbol{\theta}^k$, but with the logits $\boldsymbol{z}$ scaled by a temperature $T>0$ as $\boldsymbol{y}' = \mathrm{softmax} (\boldsymbol{z} / T)$.
$\mathcal{L}_{\mathrm{KD}}$ then replaces the second term of Eq. (\ref{eq:ewc}).

These regularization-based methods require storing the previous model's weights or $\Omega^k_i$, while the parameters in a CTC model usually occupy a large space (10M+ parameters).
The data-based methods below store samples from previous datasets, but the required capacity is unnecessarily larger.

\subsection{Data-based Methods}

\noindent\textbf{Gradient Episodic Memory (GEM).}
Here we store samples from the past to calculate the gradients \cite{Lopez17-gem}.
If the current gradient $g$ increases loss on any of the past domains, it is projected to the gradient $\Tilde{g}$ with the minimum L2 distance to $g$, or
\begin{equation}
    \begin{array}{rl}
        \underset{\Tilde{g}}{\mathrm{minimize}}~ & ~\left\|g-\Tilde{g}\right\|_2^2 \\
        \mathrm{subject~to}~ & ~\langle\Tilde{g}, g_{k-1}\rangle \geq 0
    \end{array}
    \label{eq:qp}
\end{equation}
where $g$ and $g_{k-1}$ are respectively the gradients of the CTC outputs over the current dataset $D_{k}$ and the memory $D_{k-1}^{\dagger}$, and $\langle\cdot,\cdot\rangle$ is the inner product with positive values implying similar directions.
In this paper, we constrain the capacity of $D_{k-1}^{\dagger}$ to a fixed size and balance each corpus to having the same data size, so with more new tasks, some previous samples have to be dropped.
Conventionally the data preserved for GEM are sampled randomly from the previous datasets.
For the preserved data to generate gradients representing better directions for the whole dataset, we propose two data selection methods to find samples better indicating previous data distributions.

\noindent\textbf{Minimum Perplexity (PP).}
Since topic domains vary among datasets, we propose to train an LM (RNN-LM or n-gram-LM) for each dataset $D_k$, based on which utterances with minimum perplexity are saved in the memory, assuming they better represent the linguistic property of the corpus.

\noindent\textbf{Median Length (Len).}
We noted the averaged utterance length in each dataset varies, longer in books while shorter in daily and spontaneous conversations.
Since longer and shorter utterances have slightly different acoustic properties, we propose to preserve samples with lengths close to the median.

\section{Experiments}
\label{sec:exp}

\begin{table*}[t]
	\caption{
	WERs(\%) of the CTC model without (Sec. (I)) and with (Sec. (II)) LM rescoring trained with different LLL approaches under the training order of WSJ-LS-SWB and tested on the three corpora in columns (i)(ii)(iii) and (vi)(vii)(viii).
	Columns (iv)(ix) (AVG) and (v)(x) (WERR) are respectively the average indicating the performance level and the relative WER reduction compared with fine-tuning (row (b)).
	The single corpus baseline, fine-tune baseline, and the multitask upper bound are respectively in rows (a), (b) and (i).
	}
	\label{tab:ctc-lll}
	\centering
    \vspace{-5pt}
    \begin{tabular}{l|ccc|c|c||ccc|c|c}
        \toprule
        \multirow{2}{*}{\textbf{Method}} & \multicolumn{5}{c||}{\textbf{(I) CTC}} & \multicolumn{5}{c}{\textbf{(II) CTC + RNN-LM}} \\
        \cmidrule{2-11}
         & \scriptsize{\textbf{(i) WSJ}} & \scriptsize{\textbf{(ii) LS}} & \scriptsize{\textbf{(iii) SWB}} & \scriptsize{\textbf{(iv) AVG}} & \scriptsize{\textbf{(v) WERR}} & \scriptsize{\textbf{(vi) WSJ}} & \scriptsize{\textbf{(vii) LS}} & \scriptsize{\textbf{(viii) SWB}} & \scriptsize{\textbf{(ix) AVG}} & \scriptsize{\textbf{(x) WERR}} \\
        \midrule
        \textbf{Baseline} &&&&&&&&&& \\
        ~~~(a) Single & 14.2 & 13.7 & 28.7 & 18.9 & $-$ & 11.8 & 10.8 & 23.0 & 15.2 & $-$ \\
        ~~~(b) Fine-tune & 25.1 & 38.8 & 28.8 & 30.9 & $-$ & 18.9 & 31.4 & 23.7 & 24.7 & $-$ \\
        \midrule
        \multicolumn{2}{l}{\textbf{Regularization-based}} &&&&&&&&& \\
        ~~~(c) EWC & 25.1 & 39.3 & 30.2 & 31.6 & $-$2.3\% & 19.1 & 31.8 & 24.7 & 25.2 & $-$2.0\% \\
        ~~~(d) SI & 21.9 & 32.0 & 35.7 & 29.9 & 3.2\% & 15.8 & 23.5 & 28.6 & 22.6 & 8.5\% \\
        ~~~(e) KD & 22.7 & 33.1 & 29.4 & 28.4 & 8.1\% & 16.7 & 25.4 & 24.3 & 22.2 & 10.1\% \\
        \midrule
        \textbf{Data-based} &&&&&&&&&& \\
        ~~~(f) GEM & 23.6 & 28.2 & 30.4 & 27.4 & 11.3\% & 17.1 & 21.9 & 24.8 & 21.3 & 13.8\% \\
        ~~~(g) GEM + PP & 22.8 & 27.7 & 30.3 & 26.9 & 13.3\% & 17.1 & 21.4 & 24.8 & 21.1 & 14.6\% \\
        ~~~(h) GEM + Len & 22.4 & 27.8 & 30.1 & \textbf{26.8} & \textbf{13.3\%} & 16.7 & 21.6 & 24.7 & \textbf{21.0} & \textbf{15.0\%} \\
        \midrule
        \textbf{Upper Bound} &&&&&&&&&& \\
        ~~~(i) Multitask & 10.3 & 14.1 & 25.7 & 16.7 & 46.0\% & 8.4 & 11.0 & 20.7 & 13.4 & 45.7\% \\
		\bottomrule
	\end{tabular}
    \vspace{-5pt}
\end{table*}

\subsection{Datasets}
    \label{subsec:corpus}
    
    We chose three corpora with different acoustic and topic domains to form a sequence of tasks for the ASR models to learn.
    
    \noindent\textbf{Wall Street Journal (WSJ) \cite{Paul92-WSJ}.}
    We used the si-284 set as one of the training sets and the eval92 set for evaluation.
    
    \noindent\textbf{LibriSpeech (LS) \cite{Panayotov15-libri}.}
    We used the 100-hour clean set as one of the training sets and the clean testing set for evaluation.
    
    \noindent\textbf{Switchboard (SWB) \cite{Godfrey92-swb}.}
    We chose the 300-hour LDC97S62 subset as one of the training sets and the Hub5-2000 subset for evaluation, more spontaneous and noisy compared to WSJ and LS.
    We followed the Kaldi \cite{Povey11-kaldi} "s5c" recipe to process SWB.

\subsection{Model}
    In this paper, the CTC model \cite{Graves14-ctcasr} was considered for E2E ASR.
    During evaluation, the transcription of a given utterance was either directly decoded from the CTC output distribution or with beam decoding with an additional LM.
    We considered the LLL for CTC only but not for LM since text data are easier to collect than transcribed speech.
    The training targets of CTC and LMs were both BPE subwords \cite{Sennrich16-subword} of size 256 trained on the 800M-word LM corpus from LibriSpeech.
    
    \noindent\textbf{CTC Model.}
    The CTC model \cite{Graves14-ctcasr} was composed of a 2-layer CNN for downsampling and a 5-layer BLSTM of 512 units per direction.
    We extracted 80-dimensional Mel filterbank features with delta, delta-delta and normalization.
    The sample rate of SWB is 8kHz, lower than the other two corpora; we thus upsampled all data to 16kHz.
    SpecAugment \cite{Park19-specaug} and speed perturbation \cite{Ko15-speed-perturb} were performed in all experiments.
    
    \noindent\textbf{Language Model.}
    The RNN-LM was a 2-layer LSTM of 512 units, trained with all text data from the three datasets.
    
    \noindent\textbf{Single Task Results.}
    Row (a) of Table \ref{tab:ctc-lll} lists the results as references for our CTC trained and tested on every single task without (Sec. (I)) and with (Sec. (II)) LM.
    These were the best results after trying several models and output units to balance the performance for the three tasks.
    Although different from the state-of-the-art, these WERs showed that our models worked properly with all three datasets.

\subsection{Lifelong Learning with CTC}
    \label{subsec:exp_am}
    
    We then trained the CTC model in the order of WSJ-LS-SWB (the dataset size increased and the data got more spontaneous and noisy incrementally) and then tested on the three individual corpora.
    The results are listed in Table \ref{tab:ctc-lll}.
    
    \subsubsection{Fine-tuning Baseline and Multitask Upper Bound}
    
    We set a baseline in row (b) of Table \ref{tab:ctc-lll} by fine-tuning the models successively stage by stage on the three corpora without doing anything more.
    Now we focus on CTC model without LM (Sec. (I)).
    We found results on SWB very similar (column (iii), rows (a) v.s. (b)), showing the previous two tasks provided no gain for the new domain.
    Results on WSJ and LS were seriously degraded after training with different tasks (columns (i)(ii), rows (b) v.s. (a)), which is an evidence of catastrophic forgetting.
    
    A multitask learning upper bound was trained using all the three corpora jointly and simultaneously with results listed in the bottom row (i) of Table \ref{tab:ctc-lll}.
    This was expected to offer the best performance \cite{Nguyen19-crossdomain}, since the LLL scenario assumes only minimal past data can be reused.
    Comparing to row (b) of Table \ref{tab:ctc-lll}, the multitask learning improved the ASR performance on WSJ and SWB (columns (i)(iii), row (i) of Table \ref{tab:ctc-lll}), while slightly degraded on LS  (column (ii), row (i) of Table \ref{tab:ctc-lll}), showing that training E2E ASR with multiple datasets of very different distributions might not benefit to all domains, probably because the ASR model's capacity was too small to generalize across many very different domains simultaneously.
    Still, multitask learning provided a good upper bound here otherwise.

    \subsubsection{Regularization-based Methods}

    We now inspect the results for regularization-based methods for CTC only without LM in Sec. (I) of Table \ref{tab:ctc-lll}.
    The methods EWC and SI (rows (c)(d)) both used relatively rigid constraints to limit each model parameter from drifting too far (Eq. (\ref{eq:ewc})), except with different weights $\Omega^k_i$.
    Compared to the trivial fine-tuning (row (b)), EWC offered the same or worse performance in all datasets compared to fine-tuning (rows (c) v.s. (b)).
    SI provided improvements for the first two datasets (columns (i)(ii) of row (d)) but failed to learn the last corpus (column (iii) of row (d)).
    The regularization methods here required model parameters to be close to the model for the first task (WSJ) but unable to help the model adapt to later tasks (LS and SWB).
    The scaling parameter $\lambda$ in Eq. (\ref{eq:ewc}) may play some role here, but we found it challenging to tune and never successful.
    
    In contrast, KD (row (e)) performed the best among the three regularization-based methods, reducing the catastrophic forgetting for both WSJ and LS (columns (i)(ii)) and offering good performance for the last task SWB (column (iii)).
    Also, giving a very good average or performance level considering both earlier and later tasks, significantly better than the fine-tuning (row (b)) baseline.
    The results were obviously due to the different concepts for regularization.
    Instead of constraining each parameter from drifting too far, KD tried to constrain the model output distributions close to the earlier models by KL divergence.
    This method offered much more flexibility for the model parameters to drift freely while learning sequentially.

    \subsubsection{Data-based Methods}
    
    For data-based methods (GEM) (rows (f)(g)(h)), we allowed additional memory of 50MB, corresponding to roughly 30 minutes of audio data for 16bit 16kHz files and less than 1\% of each corpus.
    Note that the regularization-based methods' storage space was equivalent to the size of the CTC model, or 406MB here, so GEM required much less memory.
    
    The results of GEM in row (f) of Table \ref{tab:ctc-lll} outperformed all regularization-based methods, including the best one KD (row (e)), showing that storing a small dataset from the past and a good concept of learning from past data was useful.
    This is probably because learning from the past data offered more freedom for the model parameters to learn across the varying tasks.
    Compared to KD, the GEM-trained model reduced the catastrophic forgetting better on LS (column (ii), rows (f) v.s. (e)) while slightly worse on WSJ and SWB (columns (i)(iii)).
    This phenomenon is consistent with the previous finding that GEM was suitable for the earlier tasks but relatively weak for the most recent task, i.e., better backward transfer \cite{Lopez17-gem}.
    The higher freedom in shifting model parameters from stage to stage may end up with less precise parameters for later tasks or a kind of trade-off between earlier and later tasks.

    Moreover, the proposed data selection PP and Len were similarly helpful and offered decent improvements over plain GEM (rows (g)(h) v.s. (f)).
    With GEM + Len, a relative reduction of 13.3\% in averaged WER was obtained compared to the fine-tuning baseline (rows (h) v.s. (b)).

    \subsubsection{With Language Model Rescoring}
    \label{subsubsec:exp_lm}
    
    A very similar trend as in Sec. (I) of Table \ref{tab:ctc-lll} can be observed in Sec. (II) when an additional LM trained with all text data from the three corpora was applied for rescoring, demonstrating the achievable performance in our scenario.
    All WERR became higher (columns (x) v.s. (v)), showing that the multi-domain LM benefited ASR decoding on all three topic domains, and the performance gap among the three corpora was narrowed.
    
    \subsection{Learning Curve}
    \begin{figure}[t]
	\centering
	\includegraphics[width=0.98\linewidth]{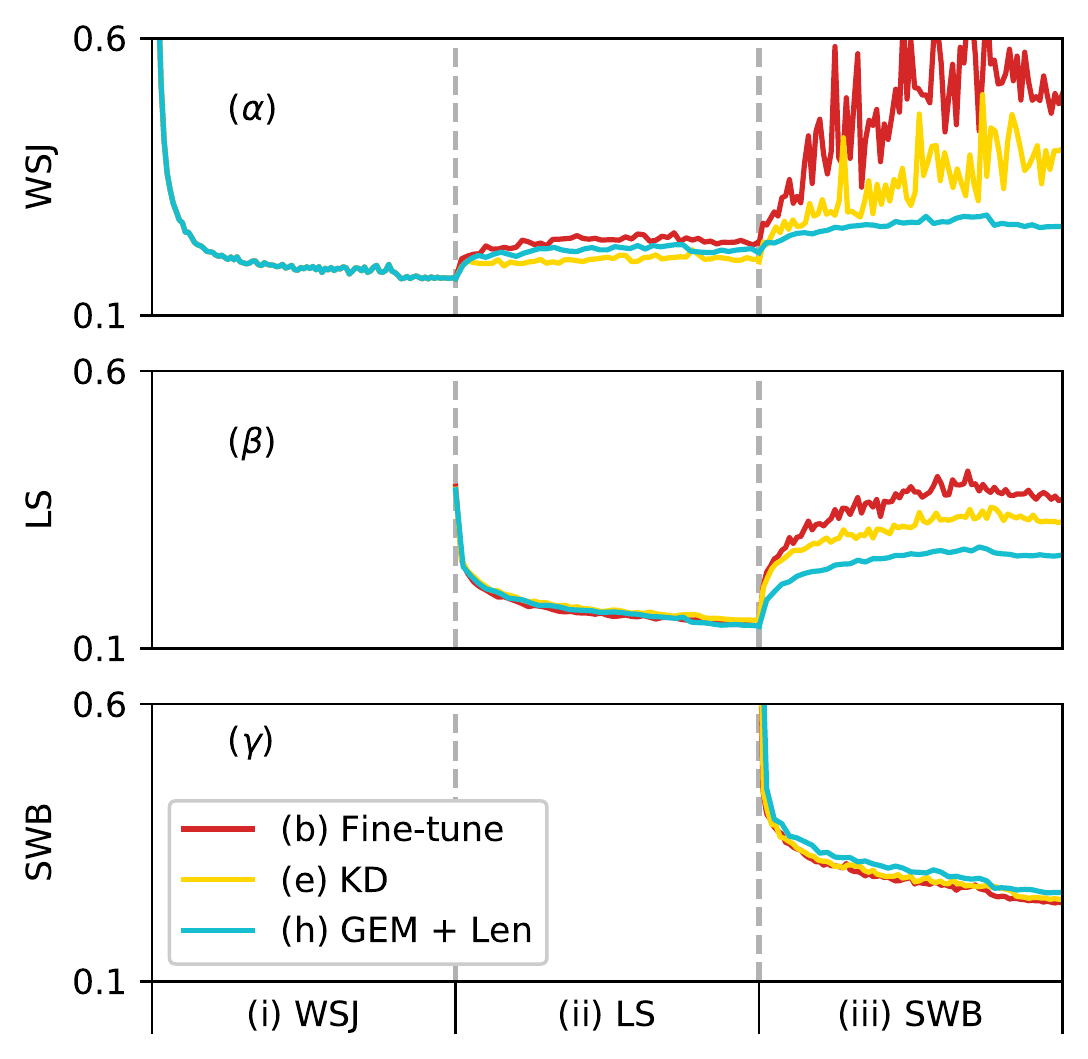}
	\caption{
	Learning curves for WERs of the CTC model under the training order of WSJ-LS-SWB, tested on ($\alpha$) WSJ, ($\beta$) LS, and ($\gamma$) SWB.
	}
	\label{fig:train_curve}
\end{figure}

    The learning curves (in WERs) of CTC models without RNN-LM, under the training order of WSJ-LS-SWB, are plotted in Fig. \ref{fig:train_curve}.
    We compared the best regularization- and data-based KD and GEM + Len approaches with the fine-tuning baseline in (rows (e)(h) v.s (b) of Table \ref{tab:ctc-lll}).
    The performance on WSJ, LS, and SWB of the CTC models during training are respectively shown in Sec. ($\alpha$)($\beta$)($\gamma$) of Fig. \ref{fig:train_curve}.
    The horizontal scale is the training steps, with each stage normalized to the same width.
    
    First, from Fig. \ref{fig:train_curve}($\alpha$), models were trained and tested on WSJ in the first stage, nothing happened, and the three curves merged into one.
    In the second and third stages of training on LS and SWB, however, all the three curves jumped up when switching the corpora and then tend to converge at higher levels, showing the phenomena of catastrophic forgetting.
    Inspecting the curves of the three methods, we found curves (e)(h) are significantly lower than curve (b), verifying KD and GEM + Len worked successfully here.
    In the third stage training with SWB, the results indicate that GEM + Len is much better than KD (curves (h) v.s. (e)).
    Moreover, curve (e) has a remarkably smaller amplitude of oscillation in the third stage, showing that exploiting a small amount of data from previous corpora stabilizes ASR training.
    Similar observations can be made in Fig. \ref{fig:train_curve}($\beta$) tested on LS, where we start to record the learning process in the second stage of training on LS.
    For the last stage in Fig. \ref{fig:train_curve}($\gamma$) trained and tested on SWB, GEM + Len performed slightly worse than the other two methods (curves (h) v.s. (b)(e)), consistent with the discussions on row (h) of Table \ref{tab:ctc-lll}, that is, GEM + Len is relatively weak for the most recent task.

    \subsection{Saved Data Size and Data Selection}
    \label{subsec:exp-data-select}
    \begin{figure}[t]
	\centering
	\includegraphics[width=0.96\linewidth]{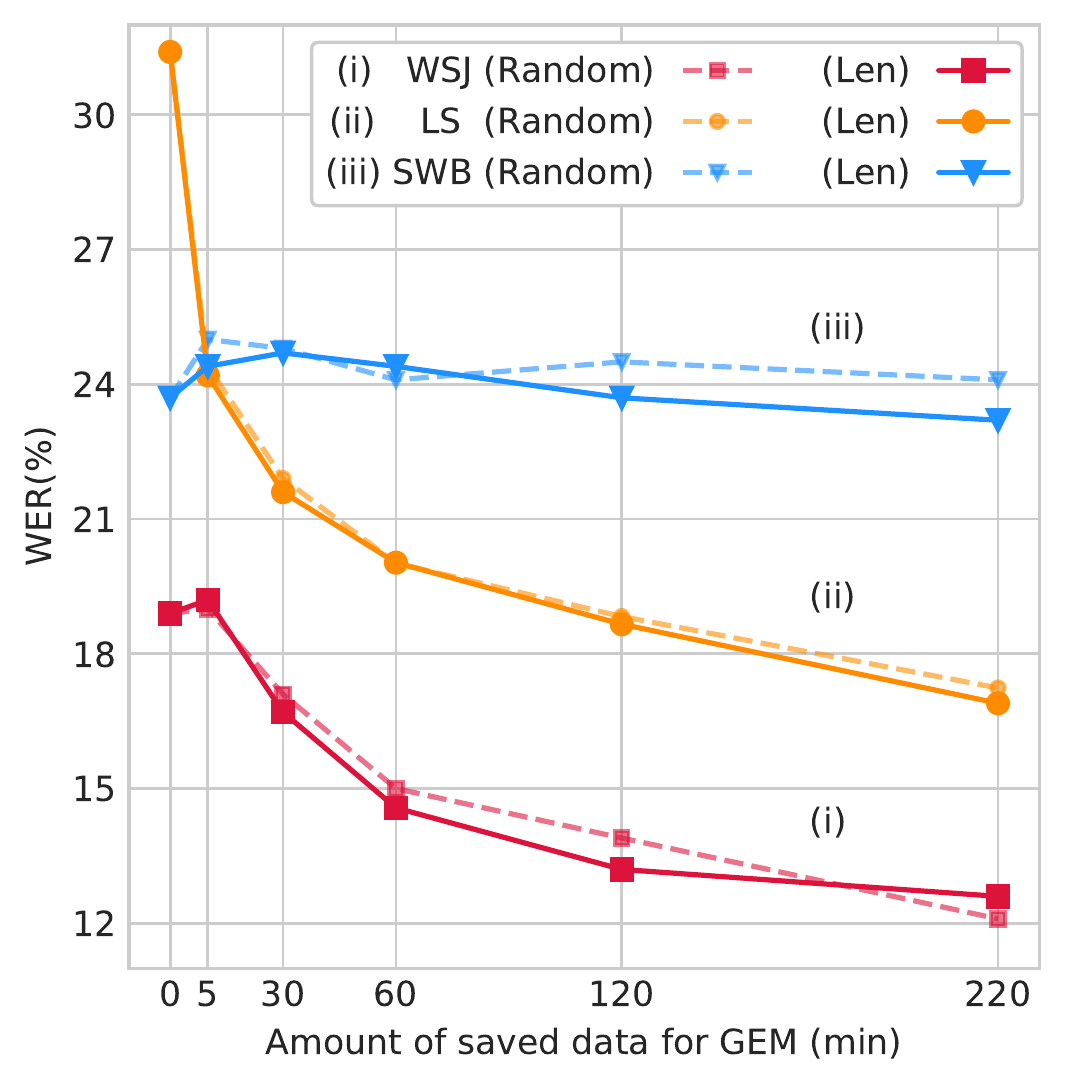}
	\caption{
	WERs with GEM under different saved data sizes, with random selection (dotted curves) or proposed Len (solid curves).
	Zero minutes of saved data is the fine-tuning baseline.
	}
	\label{fig:saved_data}
\end{figure}

    Here we investigated the effect of different saved data sizes for GEM with randomly selected saved data and GEM + Len (rows (f)(h) of Table \ref{tab:ctc-lll}), all with LM applied.
    The results in Table \ref{tab:ctc-lll} were for 30 minutes of saved data.
    We also tested with 5, 60, 120, and 220 minutes of saved data with the same setup, where 220 minutes of audio data is equivalent to the size of our CTC model.
    Results are shown in Fig. \ref{fig:saved_data}, where the leftmost points are the fine-tuning baseline (row (b), Sec. (II) of Table \ref{tab:ctc-lll}).
    
    The general trend of improved performance with increased saved data size is clear, although not very apparent for SWB, showing more saved data lead to better backward transfer.
    The averaged WER over the three tasks for GEM + Len achieved was 17.6\% WER (not shown in the figure) if 220 minutes of saved data were allowed, or a relative reduction of 28.7\% than the fine-tuning baseline.
    Yet recognizing SWB is a difficult task, GEM + Len performed slightly better than fine-tuning when 220 minutes of past data were available.
    We found 5 minutes of saved data seemed insufficient for WSJ, however, both random selection and Len offered significant improvements for LS (orange curves).
    Moreover, in most cases, the proposed Len (solid curves) excelled the random selection (dotted curves), indicating an efficient data selection algorithm is attractive.

\section{Conclusion}
\label{sec:conclusion}

This is the first paper extensively exploring the feasibility and achievable performance of LLL for E2E ASR.
We found data-based approaches were better, and proposed to properly select data from the past datasets by at least perplexity or utterance duration for mitigating catastrophic forgetting.
The proposed methods can be easily applied to other E2E ASR frameworks like LAS or RNN-T.
This is a small step towards the long term goal of having ASR technologies learning from the ever-changing real world incrementally.

\section{Acknowledgements}
We acknowledge the support of Salesforce Research Deep Learning Grant.

\bibliographystyle{IEEEtran}
\bibliography{refs}

\end{document}